\documentclass[11pt]{article}
\usepackage{coling2018}
\usepackage{times}
\usepackage{url}
\usepackage{latexsym}
\usepackage{makecell}
\usepackage{CJKutf8}
\usepackage{todonotes}
\usepackage{multirow}
 \newcommand{\squeezeup}{\vspace{-2.5mm}}

\usepackage{floatrow}
\newfloatcommand{capbtabbox}{table}[][\FBwidth]

\title{Linguistic Characteristics of Censorable Language on SinaWeibo}

\author{
Kei Yin Ng \enspace  Anna Feldman \enspace Jing Peng \enspace Chris Leberknight\\ 
Montclair State University \\
  Montclair, New Jersey, USA\\
 {\tt \{ngk2,feldmana,pengj,leberknightc\}@montclair.edu}\\
}

\date{}

\begin{document}
\maketitle
 \begin{abstract}
  This paper investigates censorship from a linguistic perspective. We collect  a corpus of censored and uncensored posts on a number of topics, build a classifier that predicts censorship decisions independent of discussion topics. Our investigation reveals that the strongest linguistic indicator of censored content of our corpus is its readability.
      
\end{abstract}

\section{Introduction}
  More than half of the world's Internet users are still restricted to a censored World Wide Web\footnote{freedomhouse.org} and do not have access to a lot of public information on the Internet.  This work is concerned with censorship directed at voices critical of the ruling government, which is common in authoritarian and repressive regimes. Our study takes a closer look at the censorship activities in mainland China, with a particular  focus on one of its mainstream social media platforms -- Sina Weibo. Internet censorship in mainland China consists of several layers:  restricted access to certain websites, restricted access to certain keyword search results, and removal of certain information published by Internet users. Since censorship on social media typically happens after a user has successfully published on the platform, what gets censored or not  is largely a ``real-time" decision due to the unpredictable nature of published content. Discussions on sensitive topics do not always get censored, as evidenced by their accessibility on the platform. So what determines a sensitive discussion to stay or to be censored? This study investigates the factors that contribute to censorship on Sina Weibo from a linguistic perspective. We locate sources that provide censored and uncensored Weibo texts, extract linguistic features from the corpus we collect and build a classifier that predicts censorship decisions independent of discussion topics. We think that identifying linguistic properties of censored content, regardless of topic, will help develop techniques to circumvent censorship.
 
  \section{Related Work}

Many measurement and circumvention studies focus more on exploiting technological limitations with existing routing protocols \cite{leberknight-etal:2012a,katti-etal:2005,levin-etal:2015,mcpherson-etal:2016,weinberg-etal:2012}).  However, little attention has focused on \textit{linguistically}-inspired techniques to study online censorship. One notable exception applies linguistic steganography to obfuscate censored content \cite{safaka-etal:2016}. Their results focus purely on circumvention while this research takes a linguistic approach to detect censorable content. 
  In recent years, several detection mechanisms have been proposed to observe and categorize the type of content and keywords that are censored \cite{knockel-etal:2015,zhu-etal:2013}.  King et al.~\shortcite{king-etal:2013} analyze the content of censored and uncensored texts from various Chinese social media sources to study the relationship between criticism of the state and chance of censorship. Their main findings suggest that negative comments about the state do not always lead to censorship. Rather, the presence of Collective Action Potential (the potential to cause collective action in real life) is what makes a post susceptible to censorship.  Lee~\shortcite{lee:2016} explores the effectiveness of linguistic tactics in circumventing online censorship in China and argued that using parodic satire could most likely survive censorship due to the nature of parodic satire -- indirect and relies heavily on users' and censors' ability to interpret based on context.  Hiruncharoenvate et al.~\shortcite{hirun-etal:2015}  discover  linguistically-informed ways to delay the detection and removal of content on Chinese social media. Their findings show that the use of homophones of censored keywords on Sina Weibo could help extend the time a Weibo post could remain available online. However, the methods of King et al.~\shortcite{king-etal:2013}, Lee~\shortcite{lee:2016} and Hiruncharoenvate et al.~\shortcite{hirun-etal:2015} all rely on a significant amount of human effort to interpret and annotate texts to evaluate the likeliness of censorship, which might not be practical to carry out for common Internet users in real life. 
Bamman et al. \shortcite{bamman-etal:2012} uncover a set of politically sensitive keywords and find that the presence of some of them in a Weibo blogpost contribute to higher chance of the post being censored. Our approach is not quite the same as Bamman et al. We target a set of topics that have been suggested to be sensitive and cover areas not limited to politics. We then study the linguistic components of both the censored and uncensored posts on those topics. In this way, we are looking beyond the mere existence of sensitive keywords in contributing to censorship. We are trying to investigate how the textual content as a whole might be relevant to censorship decision when both the censored and uncensored blogposts include the same sensitive keyword(s). 
 \section{Hypotheses}
Our experiments are based on a number of ideas from the field of psychology. 


\paragraph{1. Uncensored content is easier for readers to process than censored content} Research in psychology (e.g., Lewandowsky et al.\shortcite{lewandowsky-etal:2012}) claims that rejecting information requires cognitive effort whereas accepting information as truth is easier because detecting false information requires additional motivational and cognitive resources.  Another point of view in psychology (e.g. Schwartz et al. \shortcite{schwartz-etal:2008}) suggests that people tend to make judgments based on their subjective feelings of how easy it is to recall or process information. 
According to this group of psychologists, people do not have pre-existing, stable attitudes on many issues. Instead, their opinions are constructed on the spot and contexts (mood, emotion, ease of processing, etc.) determines the influence of the material they read and see.
Since censorship manipulates information availability, we consider it an agent that creates misinformed online experiences. By eliminating certain content, the censors are trying to present events and opinions in a biased way. 
 Based on the claims made by the research mentioned above, we hypothesize that uncensored content is easier for readers to process (or maybe even to believe) than censored content. The features described in the following sections try to address this assumption by showing that 
texts that have high readability scores (i.e., easy to process) tend to be left uncensored. We explore the psychological meaning of words and develop new readability metrics to test this hypothesis.
\paragraph{2. Censored and uncensored texts should have their own unique linguistic characteristics.}

Sali et al. \shortcite{sali-etal:2016} claim that we all have tendency to choose information that we are more frequently exposed to. We hypothesize that censored and uncensored blogposts, regardless of topics, will have their own unique recurrent themes and characteristics.

 \section{Data collection}
We collect a corpus of censored and uncensored texts in order to analyze and compare the linguistic signals embedded in each category\footnote{https://github.com/bondfeld/Datasets}. 

 
\squeezeup
\begin{table}[htp]
\begin{small}
\centering
\renewcommand{\arraystretch}{0.8}
\begin{tabular}{c|c|c|c|c}
\thead{Issue} & \thead{Date} & \thead{\makecell{Search \\Term(s)}} & \thead{\makecell{Censored \\Quantity}} & \thead{\makecell{Uncensored \\Quantity}}\\
\hline
\hline
\makecell{air \\pollution} & \makecell{3/2013 \\- 12/2017} & \makecell{smog \\(\begin{CJK*}{UTF8}{gbsn}雾霾\end{CJK*})} & 223 & 240\\
 \hline
gutter oil & \makecell{4/2012 \\- 9/2017} & \makecell{gutter oil \\(\begin{CJK*}{UTF8}{gbsn}地沟油\end{CJK*})} & 120 & 138\\
 \hline
\makecell{milk \\scandal} & \makecell{4/2012 \\- 7/2017} & \makecell{melamine\\(\begin{CJK*}{UTF8}{gbsn}三聚氰胺\end{CJK*}) \\ \& \\toxic formula\\(\begin{CJK*}{UTF8}{gbsn}毒奶粉\end{CJK*})} & 49 & 85\\
 \hline
\makecell{Internet \\censorship} & \makecell{9/2013 \\- 11/2017} & \makecell{censor\\(\begin{CJK*}{UTF8}{gbsn}屏蔽\end{CJK*})} &195 & 216\\
 \hline
\makecell{Internet \\propaganda} & \makecell{7/2014 \\- 12/2017} & \makecell{fifty-cent\\(\begin{CJK*}{UTF8}{gbsn}五毛\end{CJK*})} & 258 & 290\\
  \hline
 Bo Xilai & \makecell{1/2015 \\- 11/2017} & \makecell{Bo  Xilai \\(\begin{CJK*}{UTF8}{gbsn}薄熙来\end{CJK*})} & 125 & 75\\
 \hline
 \makecell{kindergarten \\abuse} & \makecell{11/2017 \\- 12/2017} & \makecell{RYB \\(\begin{CJK*}{UTF8}{gbsn}红黄蓝\end{CJK*})} & 53 & 94\\
 
\end{tabular}
 \caption{Published date range, search term(s) used, and quantity of each issue}\label{tbl:data}
 \end{small}
  \end{table}

\subsection{The Corpus}
Our corpus contains censored and uncensored microblogposts on scandalous issues happened or happening in mainland China. The selection of these issues is inspired by a set of lexicon, namely the Grass-mud Horse Lexicon, devised and widely used by Chinese Internet users as a political satire. Past studies have shown the relevance and significance of this lexicon in facilitating online political and social discussion among Chinese Internet users (Wang \shortcite{wang:2012}, Tang and Yang \shortcite{tang:2011}). Therefore, the lexicon can be regarded as an indicator of scandalous issues in the mainland Chinese society. Most issues covered in our data are described in the lexicon and the relevant blogposts are retrievable using relevant search terms. Table \ref{tbl:data} shows the quantity and published date range of blogposts of each issue, and also the search keyword used to obtain the blogposts. The issues are divided into four categories as below.
 
\subsubsection{Pollution and Food Safety}
This category consists of blogposts relevant to three scandalous incidents. The first is about the use of \emph{Gutter Oil} in mainland China. It was reported in 2010 that a significant number of restaurants used gutter oil as cooking oil. Long-term consumption of gutter oil is believed to cause stomach and liver cancer. The second issue is the \emph{Milk Scandal }in 2008 when milk and infant formula made in mainland China were found to be adulterated with melamine to make the products appeared to have high protein content. This caused over 50,000 infants to be hospitalized. The third issue is the ongoing severe \emph{Air Pollution} problem in some cities in mainland China. 

\subsubsection{Internet Censorship and Propaganda}
This category consists of blogposts that discuss two issues - the Internet censorship and the Internet propaganda activities on Chinese social media platforms and websites. For the propaganda issue, we target blogposts that talk about the "fifty-cent party"(\begin{CJK*}{UTF8}{gbsn}五毛党\end{CJK*}), a group of commentators that is believed to be hired by the Chinese authorities to manipulate public opinion in favor of the Chinese Communist Party.

\subsubsection{Bo Xilai}
This category is about \emph{Bo Xilai}(\begin{CJK*}{UTF8}{gbsn}薄熙来\end{CJK*}), a former Communist Party chief in Chongqing. In 2013, he was found guilty of corruption and was expelled from the Communist Party, parliament and sentenced to life imprisonment. 
\subsubsection{Kindergarten Abuse}
This category is about a recent case of \emph{Kindergarten Abuse} in Beijing. In November, 2017, toddlers at a nursery called RYB(\begin{CJK*}{UTF8}{gbsn}红黄蓝\end{CJK*}) were reported to be molested, spiked with needles and drugged with mysterious white pills.
    
\subsection{Sources}
\subsubsection{Uncensored Data}
All uncensored blogposts are collected from Sina Weibo\footnote{https://www.weibo.com/}. Sina Weibo is regarded as one of the most popular social media platforms in mainland China. It functions similarly as Twitter where users can publish, reblog and repost opinions and news on any topic. Although users can publish freely on Weibo, the published content is subject to scrutiny and would possibly be censored or deleted if it is considered to have violated Weibo's policies. Therefore, content that can be found on Sina Weibo has already passed the censorship mechanisms and is regarded as uncensored.

\subsubsection{Censored Data}
All censored data are collected from Freeweibo\footnote{https://freeweibo.com} and WeiboScope\footnote{http://weiboscope.jmsc.hku.hk}. Both sources tracked censored blogposts from Sina Weibo and make them available to the public.
 Below are some examples of censored and uncensored posts with their English translations:
 
\bigskip
\begin{small}
 \noindent
 \underline{Censored - Bo Xilai}:
 \begin{CJK*}{UTF8}{gbsn} 薄熙来是今天的高岗，周永康是今天的康生，两位都是为党和国家立下汗马功劳的人，
 也都是被政治斗争构陷的人。 \end{CJK*}\newline
 \noindent Bo Xilai is today's Gao Gang. Zhou Yongkang is today's Kang Sheng. They both made great contributions to the Party and the country. They both got framed by political infighting.



 \noindent
 \underline{Censored - Kindergarten}:
 \begin{CJK*}{UTF8}{gbsn} 每次看到这样的新闻，我都宁愿这件事情的结局是有人造谣，然后我因为传播谣言被逮进去蹲几天也不希望是真的。到底能不能来个解释啊真是日狗了！\end{CJK*}
 \noindent 
 Every time I read this kind of news, I'd rather believe they're just made-up stories. I'd rather get arrested and sent to jail for a couple of days for spreading untrue rumors than knowing the stories are real. Can anyone just give an explanation? This is really frustrating!


 
\bigskip
 \noindent
 \underline{Uncensored Bo Xilai}:
 \begin{CJK*}{UTF8}{gbsn} 薄熙来始终不认罪，戴械具，老周认罪，不戴，好看点。\end{CJK*}
 \noindent
 Bo Xilai still isn't pleading guilty. And he's cuffed. Old Zhao pleaded guilty, and he's not cuffed. That looked better.  

\noindent
 \underline{Uncensored Kindergarten}:
 \begin{CJK*}{UTF8}{gbsn} 从携程亲子园到红黄蓝幼儿园，“虐童”案再次成为焦点，只希望类似的事件从此不会发生，希望澄清的一切都是事实的真相，希望所有的小朋友都在阳光下健康成长。\end{CJK*}

 \noindent
 From Ctrip Day Care to RYB Kindergarten, "child abuse" has once again become the news highlight. I just hope nothing similar will happen again, and all the clarifications are the truth. I hope all children can grow up healthily under the sun.   

 \end{small}

\subsection{Data Selection and Preprocessing}
To ensure that a bloggpost's original text content is the only factor that renders it being censored or uncensored, only blogposts that do not contain any images, hyperlinks or reblogged content are selected.
Some blogposts returned by key-term search are not irrelevant to our target issues because of the multiple word senses or meanings of search terms. We filter out all irrelevant blogposts.
Due to the fact that the availability of censored data from the sources described above are lower (except the Bo Xilai issue), we first collect all available censored blogposts, and then based on the quantity of censored blogposts we collect corresponding number of uncensored blogposts that were published in the same date range as the censored counterpart. 
Name of author, friend tags and hashtags are removed from all data.
Finally, since the Chinese language does not have word boundaries, word segmentation has to be carried out before certain linguistic features can be extracted. We use Jieba\footnote{https://github.com/fxsjy/jieba} to segment all the data. The performance of Jeiba on our data has been manually checked to ensure quality. In cases of mis-segmentation, Jieba allows manual specification of mis-segmented words to improve accuracy.

\section{Linguistic Features}
We extract features described below for building classifiers.
\squeezeup
  \paragraph{Sensitive Keywords} \label{sec:keywords}
 We collect keywords that are regarded as sensitive in mainland China and count the frequency of keywords in each blogpost. The first source is a list of blacklisted keywords provided by Wikipedia\footnote{https://en.wikipedia.org/wiki/List\_of\_blacklisted\_keywords\_in\_China} and the second source is a list of sensitive Sina Weibo search terms provided by China Digital Times\footnote{https://chinadigitaltimes.net/china/sensitive-words-series/}. As accessibility of search results change from time to time, China Digital Times tests the ``searchability'' of each keyword and records the date of testing for reference. For each issue, we collect keywords that have been tested during the same time period as the published date of the blogposts.
 \squeezeup
\paragraph{Sentiment}
We use BaiduAI\footnote{https://ai.baidu.com} to obtain a set of sentiment scores for each blogpost. While the sentiment analyzer of BaiduAI is not designed specifically for Weibo texts, it targets customer reviews and other "comment type" of texts. We find it suitable to apply on Weibo texts which share similar characteristics with those text types. The sentiment analyzer provides a positive sentiment percentage score and a negative sentiment percentage score for each post, which sum to 1.
\squeezeup
\paragraph{LIWC}
The English Linguistic Inquiry and Word Count (LIWC) \cite{penne-etal:2007,penne-etal:2015} is a program that analyzes text on a word-by-word basis, calculating percentage of words that match each language dimension. LIWC is built on dominant theories in psychology, business, and medicine and provides word categories representing the various types of words such as personal pronouns, verbs, tenses as well as many other linguistic and psychological categories.
The Chinese LIWC dictionary is developed by Huang et al.~\shortcite{huang-etal:2012} based on the English LIWC dictionary. It is built by first translating from the English LIWC, and then further developed and modified to accommodate the linguistic differences between English and Chinese. 
We use the Chinese LIWC to extract the frequency of word categories. Altogether we extract 95 features from LIWC.


 
\squeezeup
\paragraph{Word frequency (WordFreq)}\label{sec:wordfreq} 
Inspired by Zheng~\shortcite{zheng:2005}  on assessing readability of Chinese text, we calculate the average frequency of words in each Weibo post based on Aihanyu's CNCorpus of modern Chinese\footnote{http://www.aihanyu.org/cncorpus/index.aspx}. It is a corpus that consists of about 9.5 million Chinese words. Word frequency of the corpus provides a picture of how often a certain word is used in modern Chinese texts. The lower the frequency, the less commonly used a word is. For words that appear very rarely (less than 50 times in the corpus), we count their frequency as 0.0001\%. 
\squeezeup
\paragraph{Character frequency (CharFreq)}\label{sec:charfreq}
Besides word frequency, we also extract character frequency based on the character frequency list of modern Chinese compiled by Da~\shortcite{da:2004}\footnote{http://lingua.mtsu.edu/chinese-computing/statistics/}. The idea is similar to word frequency -- the lower the character frequency, the less commonly used a character is.
\squeezeup
\paragraph{Semantic classes}\label{sec:WC/semantic classes}
Zheng~\shortcite{zheng:2005} discusses the insufficiency of relying on word frequency alone to assess text readability. The semantics of words should also be taken into consideration. The Chinese Thesaurus \begin{CJK*}{UTF8}{gbsn}同义词词林\end{CJK*} (developed by Mei~\shortcite{mei:1984} and extended by HIT-SCIR\footnote{Harbin Institute of Technology Research Center for Social Computing and Information Retrieval.}) divides words into 12 semantic classes (Human, Matter, Space and time, Abstract matter, Characteristics, Actions, Psychology, Human activities, States and phenomena, Relations, Auxiliary words, and Formulaic expressions). A word might belong to more than one semantic class. 
For example, the word \begin{CJK*}{UTF8}{gbsn}上\end{CJK*} belongs to 7 classes: Space and time, Abstract matter, Characteristics, Actions, Human activities, State and phenomena, and Relations. The more semantic classes a word belongs to, the more semantic variety it has. The idea is that more semantic variety requires more mental processing to interpret such word in relation to its surrounding context, and hence contributes to a higher difficulty level of the text. 
We count the number of distinct semantic classes found in each post, normalize by dividing the number of words by the number of semantic classes. 
   
\squeezeup
\paragraph{Readability 1}\label{sec:readability}
Inspired by Zheng~\shortcite{zheng:2005}'s discussion on the effectiveness of determining text readability with the number of distinct semantic classes a sentence has, and the insufficiency of using word frequency or word count alone to decide readability, we combine the use of frequency and semantic classes to form a readability metric for our data.   
We take the mean of character frequency, word frequency and word count to semantic groups ratio as a score of text readability. For each individual component, a lower score means a lower readability (more difficult to read and understand). Therefore, the lower the mean of the 3 components, the lower readability a text has. 
While our metric has not been formally tested, its components overlap with that of Sung et al. \shortcite{sung-et-al:2015} on the word level and semantic level. Their readability metric has been tested and validated to measure the readability of Chinese written texts for native speakers. 
\squeezeup
\paragraph{Idioms}
The part-of-speech tagger provided by Jieba annotates idioms. Most Chinese idioms are phrases that consist of 4 characters. The idiom outputs are manually checked to ensure tagging accuracy. In total, 399 and 259 idioms are extracted from censored and uncensored blogposts respectively. We normalize the raw count of idioms by the number of words in the blogpost. To see whether the presence of idioms would potentially affect the readability of a text, we calculate the correlation coefficient between the readability score (see Section \ref{sec:readability}) and the number of idioms divided by word count. We find a negative correlation between the two, suggesting that the higher the number of idioms in a text, the lower its readability. 
\squeezeup
\paragraph{Readability 2}
To further investigate how idioms might interact with the readability feature, we incorporate the inverse of normalized idioms into our readability metrics to create a second version of the readability score. Both versions are used as features for classification. In cases where the normalized idiom is 0, we re-scale it to 0.01.
\squeezeup
\paragraph{Word embeddings and eigenfeatures}\label{sec:eigen}
We use distributed word representations (word embeddings) as features. We train word vectors using the word2vec tool \cite{mikolov1,mikolov2} on 300,000 of the latest Chinese articles\footnote{ https://dumps.wikimedia.org/zhwiki/latest/} provided by Wikipedia. 
We then remove stop-words from each blogpost in our censorship corpus. For each word in the blog post, we compute a 200 dimensional vector. Since the number of words associated with each blogpost is different, we compute the 200x200 covariance matrix for each document and determine the eigen decomposition of this matrix. The eigenvectors are the directions in which the data varies the most. We use the eigenvalues as features to represent our data.
The last 40 eigenvalues capture about 85\% of total variance and are therefore used as features.

 \section{Censorship: The Human Baseline}\label{sec:human_baseline}
We carried out a crowdsourcing experiment on Amazon Mechanical Turk\footnote{https://www.mturk.com} on the Bo Xilai data. We present all blogposts one by one and ask participants to decide whether they think a post has been censored (Yes) or has not been censored (No) on Sina Weibo. To make sure the results are reliable, four control questions are included to test participants' knowledge on the Chinese language and the Bo Xilai incident (Please see Appendix 1). Only responses that correctly answer all four control questions are accepted and analyzed. The same experiment is also presented as a survey to five acquaintances of the author who are native Chinese speakers and are knowledgeable of the Bo Xilai incident and Internet censorship in mainland China. In total, we collect responses from 23 participants. Not all 23 participants answered all posts. Each post has been judged by four to eight different participants. The average accuracy of the human judges is 63.51\% which serves as our human baseline. The interannotator agreement, however, is low (Cohen's kappa \cite{cohen:60}) is 0.07), which suggests that the task of deciding what blogpost has been censored is extremely difficult.

\section{Machine Learning Classification}\label{sec:automatic}

\squeezeup
\begin{table}[th]
\begin{center}
\setlength{\tabcolsep}{.7ex}
\begin{small}
\begin{tabular}{l|c||ccc|ccc}
{\bf Features } &  & \multicolumn{3}{c|}{\bf Censored}& \multicolumn{3}{c}{\bf Uncensored}\\
\hline
& Acc & Pre & 	Rec & 	F1 & 	Pre & 	Rec & 	F1	\\
\hline
NB all (147) &0.65 &0.76 &0.65 &0.70 &0.53 &0.65 &0.58 \\
NB eigenvalues (40) &0.57 &0.69 &0.57 &0.62 &0.43 &0.57 &0.50 \\
NB ling. features (107) &0.64 &0.76 &0.61 &0.68 &0.51 &0.68 &0.58 \\
NB best features(17) & 0.67&0.74 & 0.74& 0.74 &0.56 &0.56 & 0.56 \\
\hline
SMO all (147) &0.70 &0.75 &0.78 &0.77 &0.61 &0.56 &0.58 \\
SMO eigenvalues (40) &0.62 &0.63 &0.92 &0.75 &0.44 &0.11 &0.17 \\
SMO ling. features (107)&0.68 &0.74 &0.74 &0.74 &0.57 &0.57 &0.57 \\
SMO best features (17) & 0.72& 0.71 & 0.91 &0.80 & 0.73 & 0.39 & 0.50 \\
\hline
majority class &0.63&&&&&&\\
\hline
\end{tabular}
  \caption{Classification results for the Bo Xilai subcorpus.}\label{tbl:boxilai}
\end{small}
\end{center}
\end{table}

\begin{table}[th]
\begin{center}
\setlength{\tabcolsep}{.7ex}
\begin{small}
\begin{tabular}{l|c||ccc|ccc}
{\bf Features } &  & \multicolumn{3}{c|}{\bf Censored}& \multicolumn{3}{c}{\bf Uncensored}\\
\hline
& Acc & Pre & 	Rec & 	F1 & 	Pre & 	Rec & 	F1	\\
\hline
NB all (147) &0.63 &0.49 &0.64 &0.55 &0.75 &0.62 &0.68 \\
NB eigenvalues (40) &0.58 &0.45 &0.72 &0.55 &0.76 &0.50 &0.60 \\
NB ling. features (107) &0.64 &0.50 &0.59 &0.54 &0.74 &0.67 &0.70 \\
NB best features (9) & 0.72 & 0.64 & 0.53 & 0.58 & 0.76& 0.83 & 0.72 \\
\hline
SMO all (147) &0.63 &0.47 &0.34 &0.40 &0.68 &0.79 &0.73 \\
SMO eigenvalues (40) &0.63 &0.00 &0.00 &0.00 &0.64 &0.99 &0.78 \\
SMO ling. features (107)&0.65 &0.53 &0.36 &0.43 &0.69 &0.82 &0.75 \\
SMO best features (9) &0.70 & 0.91 & 0.19 &0.31 & 0.68&0.99 & 0.81\\
\hline
majority class &0.64&&&&&&\\
\hline
\end{tabular}
\caption{Classification results for the Kindergarten Abuse subcorpus.}\label{tbl:kindergarten}
\end{small}
\end{center}
\end{table} 
 
\begin{table}[th]
\begin{center}
\setlength{\tabcolsep}{.7ex}
\begin{small}
\begin{tabular}{l|c||ccc|ccc}
{\bf Features } &  & \multicolumn{3}{c|}{\bf Censored}& \multicolumn{3}{c}{\bf Uncensored}\\
\hline
& Acc & Pre & 	Rec & 	F1 & 	Pre & 	Rec & 	F1	\\
\hline
NB all (147) &0.66 &0.60 &0.76 &0.67 &0.74 &0.57 &0.64 \\
NB eigenvalues (40) &0.62 &0.57 &0.68 &0.62 &0.68 &0.57 &0.62 \\
NB ling. features (107) &0.69 &0.63 &0.78 &0.70 &0.77 &0.61 &0.68 \\
NB best features(69) & 0.67 & 0.62 & 0.74 & 0.67 & 0.74& 0.61& 0.67 \\
\hline
SMO all (147) &0.71 &0.70 &0.66 &0.68 &0.73 &0.76 &0.74 \\
SMO eigenvalues (40) &0.60 &0.56 &0.63 &0.59 &0.65 &0.58 &0.61 \\
SMO ling. features (107)&0.72 &0.70 &0.68 &0.69 &0.74 &0.75 &0.75 \\
SMO best features (69) &0.71 & 0.69& 0.68 & 0.68 & 0.73 & 0.74 & 0.73\\
\hline
majority class &0.54&&&&&&\\
\hline
\end{tabular}
\caption{Classification results for the Pollution \& Food Safety  subcorpus.}\label{tbl:health-standard}
\end{small}
\end{center}
\end{table}
 
\begin{table}[th]
\begin{center}
\setlength{\tabcolsep}{.7ex}
\begin{small}
\begin{tabular}{l|c||ccc|ccc}
{\bf Features } &  & \multicolumn{3}{c|}{\bf Censored}& \multicolumn{3}{c}{\bf Uncensored}\\
\hline
& Acc & Pre & 	Rec & 	F1 & 	Pre & 	Rec & 	F1	\\
\hline
NB all (147) &0.66&0.60&0.76&0.67&0.74&0.57&0.64\\
NB eigenvalues (40) & 0.55& 0.52&0.65&0.58&0.60&0.47&0.52\\
NB ling. features (107) &0.64&0.62&0.62&0.62&0.66&0.66&0.66\\
NB best features(33) &0.65&0.65&0.59&0.62&0.66&0.71&0.68\\
\hline
SMO all (147) &0.66 &0.63&0.65&0.64&0.68&0.67&0.67\\
SMO eigenvalues (40) & 0.56&0.55&0.40&0.46&0.57&0.70&0.63\\
SMO ling. features (107)& 0.63&0.61& 0.61& 0.61& 0.65&0.66&0.65\\
SMO best features (33) &0.67&0.64&0.66&0.65&0.67&0.67&0.67\\
\hline
majority class &0.53&&&&&&\\
\hline
\end{tabular}
\caption{Classification results for the Internet \& Propaganda subcorpus.}\label{tbl:internet}	
\end{small}
\end{center}
\end{table}
 
 \begin{table}[th]
\begin{center}
\setlength{\tabcolsep}{.7ex}
\begin{small}
\begin{tabular}{l|c||ccc|ccc}
{\bf Features } &  & \multicolumn{3}{c|}{\bf Censored}& \multicolumn{3}{c}{\bf Uncensored}\\
\hline
& Acc & Pre & 	Rec & 	F1 & 	Pre & 	Rec & 	F1	\\
\hline
SMO (All features (148)) & 0.64 & 0.63&0.58&0.60& 0.65&0.69& 0.67\\
readability 1 (1) & 0.51 & 0.43&0.11 &0.17& 0.52&0.88&0.65\\
readability 2 (1) &0.53& 0.46& 0.02&0.04& 0.53&0.98&0.69\\
best features (82) &0.66 & 0.65 &0.56&0.61&0.65&0.73&0.69\\
Majority Class &0.53 &&&&&&\\

\hline
\end{tabular}
\caption{Linear SVM Classification results for the censorship corpus. All sub-corpora combined together.}\label{tbl:all}	
\end{small}
\end{center}
\end{table}

\subsection{Naive Bayes and Support Vector Machine}
 We extract 107 linguistic features described above and 40 eigenvalues (as they capture around 85\% of total variance, see section \ref{sec:eigen}) for each blogpost. The 107 linguistic features for each blogpost are first standardized as a z-score (the standard score) before proceeding to classification. 
Then, for each of the 4 sub-corpora, we build a standard Naive Bayes \cite{John1995} classifier and a linear Support Vector Machine (SVM with SMO) \cite{platt:98} classifier (both with 10-fold cross-validation) using separately the eigenvalues only, the linguistic features only, and the combination of both (denoted as \textit{all}) (see Tables \ref{tbl:boxilai}--\ref{tbl:internet}).

\subsection{Best Features}
We use the standard Information Gain (IG) feature selection algorithm \cite{peng-etal:2005} with 10-fold cross-validation to get the features that provide the most information gain with respect to class for each sub-corpus. Tables \ref{tbl:boxilai}--\ref{tbl:internet}) summarize our experiments. While the best features for each sub-corpus are different, \textit{Readability 1} (the one that does not include the idiom count) and \textit{Word Count (WC)} appear in all 4 sets of best features.  We then use the Readability 1 feature as the \textit{only} feature to get classification results. Surprisingly, the Readability 1 feature alone achieves comparable classification accuracies as using all features we extract -- 0.65, 0.65, 0.66, 0.54 respectively on the Bo Xilai, Kindergarten Abuse, Pollution \& Food Safety, and Internet Censorship \& Propaganda sub-corpus.

\subsubsection{Readability and Word Count}
The average readability score of all uncensored blogposts is higher than that of censored, i.e., uncensored blogposts on average are less difficult to read and understand.
The average word count of all uncensored blogposts is higher than that of censored (i.e. uncensored blogposts on average are longer). 
The higher readability score of uncensored blogposts goes along with our hypothesis that uncensored blogposts tend to be easier to read. The higher readability score of uncensored blogposts also suggests that censored content could be semantically less straightforward and/or use more uncommon words. As microbloggers are aware of the possibility of censorship when discussing sensitive topics, various linguistic techniques might have been adopted such as satire as suggested by Lee ~\shortcite{lee:2016} and homophones as suggested by Hiruncharoenvate et. al~\shortcite{hirun-etal:2015}. Text readability decreases as microbloggers try to evade censorship.
It is interesting to note that we do include each component of our readability metric (character frequency, word frequency and word count to semantic groups ratio) as an individual feature for the classifiers but they do not top the best features. 

While the length of a text alone does not directly indicate readability, there might be implications when a longer blogpost is posted on a social media or microblog platform, where short texts are the norm.
While a longer blogpost does not guarantee readership, it might invite more attention when users are browsing among other relatively short texts. One of the explanations why long and easy to read blogposts are deliberately left untouched by the censors is that these posts are expected to impact the readers' opinions as is predicted by  Schwarz et al. \shortcite{schwartz-etal:2008} who claim that easier materials are more influential to readers. 

\bigskip

\squeezeup  

\squeezeup  
\section{Discussion}
 
To get a better understanding of the language differences between censored and uncensored text and what words are associated with what class, we subtract the average percentage of words in each feature from the best feature set in the censored posts from their corresponding uncensored values. A positive result indicates an association between a feature and the uncensored class, whereas a negative result indicates an association between a feature and the censored class.  We find that the use of third person plural pronoun (`they'), swear words (`fuck', `damn', `shit' etc.), and words that express anger (`hate', `annoyed', `kill' etc.) and certainty (`always', `never' etc.) are more associated with censored blogposts across all 4 sub-corpora, whereas interrogatives (`how', `when', `what' etc.), the Internet slang (`lol', `plz', `thx' etc.), and words that talk about the present (`today', `now' etc.), sadness (`crying', `grief' etc.) and vision (`view', `see' etc.) are more associated with the uncensored blogposts for all 4 sub-corpora. 
This suggests that blogposts that express opinions in a furious, cursing manner and with certainty might be more susceptible to censorship, while blogposts that are more casual and appealing (the heavier use of Internet Slangs), more focused on discussing or querying the current state of matter (interrogatives, present words, and vision words), or describe negative emotion in a less intimidating way are tend to left uncensored. In a sense, by not blocking this type of posts, the censors provide an outlet for the bloggers to express their  criticism and negative emotions without growing angry, since anger might lead to action.
     
While we do not explicitly experiment on collective action potential (CAP) proposed by King et al.~\shortcite{king-etal:2013}, some of our findings show characteristics common with CAP -- social engagement. 
According to Pennebaker et al. \shortcite{penne-etal:2015}, ``third person pronouns are, by definition, markers to suggest that the speaker is socially engaged or aware." Generally, the features typical of censored content convey social engagement, confidence and anger, and involve fewer words that refer to senses, but more words that describe cognitive processes. 
This finding also goes along with our hypothesis that censored and uncensored texts each has their own unique linguistic characteristics.

Although incorporating idioms in the readability metrics does not improve the performance of the models on the merged corpus, we noticed a slight improvement in the performance of the classifier on the Pollution \& Environment and the Internet Censorship sub-corpora. The normalized idiom count are selected as one of the best features for these two datasets. It is worth noting that 39\% and 22\% of censored and uncensored blogposts respectively contain idioms overall, and that idioms are negatively correlated with the Readability 1 score. For the Pollution \& Environment and the Internet Censorship sub-corpora, the differences in the usage of idioms  between the censored and uncensored classes are more prominent. It is yet to be investigated in what other ways idioms might contribute to the prediction of censorship and how to incorporate them into the model.

\section{Limitations}
Since we compare censored and uncensored data that share the same keyword(s) in their textual content, and due to the fact that the SinaWeibo does not allow more than 200 results to be returned from their keyword search API, our data is collected manually and therefore the quantity is not large. Also, our censored data source does not necessarily track the publishing history of every single SinaWeibo user. Our goal is to further improve the representativeness of our data in future work. Although Bamman et al. \shortcite{bamman-etal:2012} has shown the geographical differences in censorship rate of blogposts, our concern is on textual content only. It is possible that some fine-grained linguistic differences might exist in censored data due to geographical differences. However, this is left for future work to gauge such patterns, if they exist.

 \section{Conclusion}
We described a pilot study in which we built a model to classify censored and uncensored social media posts from mainland China. We deliberately did not use topics as classification features because some censorable topics vary throughout time and across countries, but linguistic fingerprints should not.  Our goal is to explore whether linguistic features could be effective in discriminating between censored and uncensored content. We have found such features that characterize the two categories and built classifiers that achieve promising results.
The focus of this study is on censorship in mainland China. However, there are many other regions around the world that exercise Internet censorship. It is important to verify if the features we identified hold cross-linguistically. 
  

\section*{Acknowledgements}

We thank the anonymous reviewers for their careful reading of this article and their many insightful comments and suggestions. This work is supported by the National Science Foundation under Grant No.: 1704113, Division of Computer and Networked Systems, Secure \& Trustworthy Cyberspace (SaTC).
\newpage

\bibliography{censorship}
\bibliographystyle{acl}
\end{document}